\documentclass{article}

\usepackage{times}
\usepackage{graphicx} 
\usepackage{subfigure} 

\usepackage{natbib}

\usepackage{algorithm}
\usepackage{algorithmic}

\usepackage{amsmath}
\DeclareMathOperator{\sign}{sign}

\usepackage{hyperref}

\usepackage{xcolor}
\definecolor{codegray}{gray}{0.95}
\newcommand{\code}[1]{\colorbox{codegray}{\texttt{#1}}}

\newcommand{\foolbox}{Foolbox}
\newcommand{\x}{{\bf x}}
\newcommand{\g}{{\bf g}}

\usepackage[T1]{fontenc}
\usepackage{inconsolata}

\usepackage{bbm}

\usepackage[accepted]{icml2017}

\icmltitlerunning{Foolbox: A Python toolbox to benchmark the robustness of machine learning models}

\begin{document} 

\twocolumn[
\icmltitle{Foolbox: A Python toolbox to benchmark the \texorpdfstring{\\}{}
           robustness of machine learning models}



\icmlsetsymbol{equal}{*}

\begin{icmlauthorlist}
\icmlauthor{Jonas Rauber}{equal,cin,bccn,imprs}
\icmlauthor{Wieland Brendel}{equal,cin,bccn}
\icmlauthor{Matthias Bethge}{cin,bccn,mpibc,physik}
\end{icmlauthorlist}

\icmlaffiliation{cin}{Centre for Integrative Neuroscience, University of T\"ubingen, Germany}
\icmlaffiliation{mpibc}{Max Planck Institute for Biological Cybernetics, T\"ubingen, Germany}
\icmlaffiliation{bccn}{Bernstein Center for Computational Neuroscience, T\"ubingen, Germany}
\icmlaffiliation{imprs}{International Max Planck Research School for Intelligent Systems, T\"ubingen, Germany}
\icmlaffiliation{physik}{Institute for Theoretical Physics, University of T\"ubingen, Germany}

\icmlcorrespondingauthor{Jonas Rauber}{jonas.rauber@bethgelab.org}

\icmlkeywords{python, adversarial examples, robustness, neural networks, machine learning, adversarials, adversarial attacks}

\vskip 0.3in
]



\begin{NoHyper}
\printAffiliationsAndNotice{\icmlEqualContribution{}} 
\end{NoHyper}

\begin{abstract} 
Even todays most advanced machine learning models are easily fooled by almost imperceptible perturbations of their inputs. \textit{Foolbox} is a new Python package to generate such adversarial perturbations and to quantify and compare the robustness of machine learning models. It is build around the idea that the most comparable robustness measure is the minimum perturbation needed to craft an adversarial example. To this end, Foolbox provides reference implementations of most published adversarial attack methods alongside some new ones, all of which perform internal hyperparameter tuning to find the minimum adversarial perturbation. Additionally, Foolbox interfaces with most popular deep learning frameworks such as PyTorch, Keras, TensorFlow, Theano and MXNet and allows different adversarial criteria such as targeted misclassification and top-k misclassification as well as different distance measures. The code is licensed under the MIT license and is openly available at \url{https://github.com/bethgelab/foolbox}. The most up-to-date documentation can be found at \url{http://foolbox.readthedocs.io}.
\end{abstract}

In \citeyear{szegedy2013}, \citeauthor{szegedy2013} demonstrated that minimal perturbations, often almost imperceptible to humans, can have devastating effects on machine predictions. These so-called \textit{adversarial perturbations} thus demonstrate a striking difference between human and machine perception. As a result, adversarial perturbations have been subject to many studies concerning the generation of such perturbations and strategies to protect machine learning models such as deep neural networks against them.

A practical definition of the robustness $R$ of a model, first used by \citet{szegedy2013}, is the average size of the minimum adversarial perturbation $\rho(\x)$ across many samples $\x$,
\begin{equation}
    R = \left\langle \rho(\x)\right\rangle_\x \quad \text{where}
\end{equation}
\begin{equation}
    \rho(\x) = \min_{\boldsymbol\delta} d(\x, \x + {\boldsymbol\delta})\quad\text{s.t.}\quad   \x + {\boldsymbol\delta} \text{ is adversarial}
\end{equation}
and \( d(\cdot) \) is some distance measure.

Unfortunately, finding the global minimum adversarial perturbation is close to impossible in any practical setting, and we thus employ heuristic attacks to find a suitable approximation. Such heuristics, however, can fail, in which case we could easily be mislead to believe that a model is robust \citep{brendel17}. Our best strategy is thus to employ as many attacks as possible, and to use the minimal perturbation found across all attacks as an approximation to the true global minimum.

At the moment, however, such a strategy is severely obstructed by two problems: first, the code for most known attack methods is either not available at all, or only available for one particular deep learning framework. Second, implementations of the same attack often differ in many details and are thus not directly comparable. \foolbox{} improves upon the existing Python package \textit{cleverhans} by \citet{cleverhans} in three important aspects:
\begin{enumerate}
    \item It interfaces with most popular machine learning frameworks such as PyTorch, Keras, TensorFlow, Theano, Lasagne and MXNet and provides a straight forward way to add support for other frameworks,
    \item it provides reference implementations for more than 15 adversarial attacks with a simple and consistent API, and
    \item it supports many different criteria for adversarial examples, including custom ones.
\end{enumerate}

This technical report is structured as follows: In section 1 we provide an overview over \foolbox{} and demonstrate how to benchmark a model and report the result. In section 2 we describe the adversarial attack methods that are implemented in \foolbox{} and explain the internal hyperparameter tuning.

\section{\foolbox{} Overview}

\subsection{Structure}

Crafting adversarial examples requires five elements: first, a \textbf{model} that takes an input (e.g. an image) and makes a prediction (e.g. class-probabilities). Second, a \textbf{criterion} that defines what an adversarial is (e.g. misclassification). Third, a \textbf{distance measure} that measures the size of a perturbation (e.g. L1-norm). Finally, an \textbf{attack algorithm} that takes an input and its label as well as the model, the adversarial criterion and the distance measure to generate an \textbf{adversarial perturbation}.

The structure of \foolbox{} naturally follows this layout and implements five Python modules (models, criteria, distances, attacks, adversarial) summarized below.

\paragraph{Models}\mbox{}\\
\code{foolbox.models}\\
This module implements interfaces to several popular machine learning libraries:
\begin{itemize}
    \item TensorFlow \citep{tensorflow} \\
    \code{foolbox.models.TensorFlowModel}
    \item PyTorch \citep{pytorch} \\
    \code{foolbox.models.PyTorchModel}
    \item Theano \citep{theano} \\
    \code{foolbox.models.TheanoModel}
    \item Lasagne \citep{lasagne} \\
    \code{foolbox.models.LasagneModel}
    \item Keras (any backend) \citep{keras} \\
    \code{foolbox.models.KerasModel}
    \item MXNet \citep{mxnet} \\
    \code{foolbox.models.MXNetModel}
\end{itemize}



Each interface is initialized with a framework specific representation of the model (e.g. symbolic input and output tensors in TensorFlow or a neural network module in PyTorch). The interface provides the adversarial attack with a standardized set of methods to compute predictions and gradients for given inputs. It is straight-forward to implement interfaces for other frameworks by providing methods to calculate predictions and gradients in the specific framework.

Additionally, \foolbox{} implements a
\code{CompositeModel}
that combines the predictions of one model with the gradient of another. This makes it possible to attack non-differentiable models using gradient-based attacks and allows transfer attacks of the type described by \citet{compositeattack}.

\paragraph{Criteria}\mbox{}\\
\code{foolbox.criteria}\\
A \textit{criterion} defines under what circumstances an [input, label]-pair is considered an adversarial. The following criteria are implemented:
\begin{itemize}
    \item Misclassification \\
    \code{foolbox.criteria.Misclassification} \\
    Defines adversarials as inputs for which the predicted class is not the original class.
    \item Top-k Misclassification \\
    \code{foolbox.criteria.TopKMisclassification} \\
    Defines adversarials as inputs for which the original class is not one of the top-k predicted classes.
    \item Original Class Probability \\
    \code{foolbox.criteria.OriginalClassProbability} \\
    Defines adversarials as inputs for which the probability of the original class is below a given threshold.
    \item Targeted Misclassification \\
    \code{foolbox.criteria.TargetClass} \\
    Defines adversarials as inputs for which the predicted class is the given target class.
    \item Target Class Probability \\
    \code{foolbox.criteria.TargetClassProbability} \\
    Defines adversarials as inputs for which the probability of a given target class is above a given threshold. 
\end{itemize}
Custom adversarial criteria can be defined and employed. Some attacks are inherently specific to particular criteria and thus only work with those.


\paragraph{Distance Measures}\mbox{}\\
\code{foolbox.distances}\\
Distance measures are used to quantify the size of adversarial perturbations. \foolbox{} implements the two commonly employed distance measures and can be extended with custom ones:
\begin{itemize}
    \item Mean Squared Distance\\
    \code{foolbox.distances.MeanSquaredDistance}\\
    Calculates the mean squared error\\$d({\bf x}, {\bf y}) = \frac{1}{N} \sum_i (x_i - y_i)^2$\\between two vectors ${\bf x}$ and ${\bf y}$.
    \item Mean Absolute Distance\\
    \code{foolbox.distances.MeanAbsoluteDistance}\\
    Calculates the mean absolute error\\$d({\bf x}, {\bf y}) = \frac{1}{N} \sum_i |x_i - y_i|$\\between two vectors ${\bf x}$ and ${\bf y}$.
    \item $L\infty{}$\\
    \code{foolbox.distances.Linfinity}\\
    Calculates the $L\infty{}$-norm $d({\bf x}, {\bf y}) = \max_i |x_i - y_i|$ between two vectors ${\bf x}$ and ${\bf y}$.
    \item $L0$\\
    \code{foolbox.distances.L0}\\
    Calculates the $L0$-norm $d({\bf x}, {\bf y}) = \sum_i \mathbbm{1}_{x_i \ne y_i}$ between two vectors ${\bf x}$ and ${\bf y}$.
\end{itemize}

To achieve invariance to the scale of the input values, we normalize each element of ${\bf x, y}$ by the difference between the smallest and largest allowed value (e.g. 0 and 255).

\paragraph{Attacks}\mbox{}\\
\code{foolbox.attacks}\\
\foolbox{} implements a large number of adversarial attacks, see section 2 for an overview. Each attack takes a model for which adversarials should be found and a criterion that defines what an adversarial is. The default criterion is \textit{misclassification}. It can then be applied to a reference input to which the adversarial should be close and the corresponding label. Attacks perform internal hyperparameter tuning to find the minimum perturbation. As an example, our implementation of the fast gradient sign method (FGSM) searches for the minimum step-size that turns the input into an adversarial. As a result there is no need to specify hyperparameters for attacks like FGSM. For computational efficiency, more complex attacks with several hyperparameters only tune some of them.

\paragraph{Adversarial}\mbox{}\\
\code{foolbox.adversarial}\\
An instance of the adversarial class encapsulates all information about an adversarial, including which model, criterion and distance measure was used to find it, the original unperturbed input and its label or the size of the smallest adversarial perturbation found by the attack.

An adversarial object is automatically created whenever an attack is applied to an [input, label]-pair. By default, only the actual adversarial input is returned.
Calling the attack with \texttt{unpack} set to \texttt{False} returns the full object instead.
Such an adversarial object can then be passed to an adversarial attack instead of the [input, label]-pair, enabling advanced use cases such as pausing and resuming long-running attacks.

\subsection{Reporting Benchmark Results}
\label{benchmarking}

When reporting benchmark results generated with \foolbox{} the following information should be stated:
\vspace*{-10pt}
\begin{itemize}
\itemsep-3pt
    \item the version number of \foolbox{},
    \item the set of input samples,
    \item the set of attacks applied to the inputs,
    \item any non-default hyperparameter setting,
    \item the criterion and
    \item the distance metric.
\end{itemize}



\subsection{Versioning System}

Each release of \foolbox{} is tagged with a version number of the type MAJOR.MINOR.PATCH that follows the principles of semantic versioning\footnote{\url{http://semver.org/}} with some additional precautions for comparable benchmarking. We increment the
\begin{enumerate}
    \item MAJOR version when we make changes to the API that break compatibility with previous versions.
    \item MINOR version when we add functionality or make backwards compatible changes that can affect the benchmark results.
    \item PATCH version when we make backwards compatible bug fixes that do not affect benchmark results.
\end{enumerate}
Thus, to compare the robustness of two models it is important to use the same MAJOR.MINOR version of \foolbox{}. Accordingly, the version number of \foolbox{} should always be reported alongside the benchmark results, see section \ref{benchmarking}.

\section{Implemented Attack Methods}
\label{attacks}

We here give a short overview over each attack method implemented in \foolbox{}, referring the reader to the original references for more details. We use the following notation:
\vspace*{-\baselineskip}
\begin{table}[htbp]
\centering 
\begin{tabular}{r p{5cm} }
$\x$ & a model input\\
$\ell$ & a class label\\
$\x_0$ & reference input \\
$\ell_0$ & reference label\\
$L(\x, \ell)$ & loss (e.g. cross-entropy)\\
$[b_{\text{min}}, b_{\text{max}}]$ & input bounds (e.g. 0 and 255)
\end{tabular}
\label{tab:notation}
\end{table}
\vspace*{-20pt}

\subsection{Gradient-Based Attacks}
Gradient-based attacks linearize the loss (e.g. cross-entropy) around an input $\x$ to find directions $\boldsymbol\rho$ to which the model predictions for class $\ell$ are most sensitive to,
\begin{equation}
    L(\x + \boldsymbol\rho, \ell) \approx L(\x, \ell) + \boldsymbol\rho^\top\nabla_\x L(\x, \ell).
\end{equation}
Here $\nabla_\x L(\x, \ell)$ is referred to as the gradient of the loss w.r.t. the input $\x$.

\paragraph{Gradient Attack}\mbox{}\\
\code{foolbox.attacks.GradientAttack}\\
This attack computes the gradient $\g(\x_0) = \nabla_\x L(\x_0, \ell_0)$ once and then seeks the minimum step size $\epsilon$ such that $\x_0 + \epsilon \g(\x_0)$ is adversarial.

\paragraph{Gradient Sign Attack (FGSM)}\mbox{}\\
\code{foolbox.attacks.GradientSignAttack}\\
\code{foolbox.attacks.FGSM}\\
This attack computes the gradient $\g(\x_0) = \nabla_\x L(\x_0, \ell_0)$ once and then seeks the minimum step size $\epsilon$ such that $\x_0 + \epsilon \sign(\g(\x_0))$ is adversarial \citep{fgsm}.

\paragraph{Iterative Gradient Attack}\mbox{}\\
\code{foolbox.attacks.IterativeGradientAttack}\\
Iterative gradient ascent seeks adversarial perturbations by maximizing the loss along small steps in the gradient direction $\g(\x)=\nabla_\x L(\x, \ell_0)$, i.e. the algorithm iteratively updates $\x_{k+1} \leftarrow \x_k + \epsilon \g(\x_k)$. The step-size $\epsilon$ is tuned internally to find the minimum perturbation.

\paragraph{Iterative Gradient Sign Attack}\mbox{}\\
\code{foolbox.attacks.IterativeGradientSignAttack}\\
Similar to iterative gradient ascent, this attack seeks adversarial perturbations by maximizing the loss along small steps in the ascent direction $\sign(\g(\x)) = \sign\left(\nabla_\x L(\x, \ell_0)\right)$, i.e. the algorithm iteratively updates $\x_{k+1} \leftarrow \x_k + \epsilon \sign(\g(\x_k))$. The step-size $\epsilon$ is tuned internally to find the minimum perturbation.

\paragraph{DeepFool $L2$ Attack}\mbox{}\\
\code{foolbox.attacks.DeepFoolL2Attack}\\
In each iteration DeepFool \citep{deepfool} computes for each class $\ell\ne\ell_0$ the minimum distance $d(\ell, \ell_0)$ that it takes to reach the class boundary by approximating the model classifier with a linear classifier. It then makes a corresponding step in the direction of the class with the smallest distance.

\paragraph{DeepFool $L\infty{}$ Attack}\mbox{}\\
\code{foolbox.attacks.DeepFoolLinfinityAttack}\\
Like the DeepFool L2 Attack, but minimizes the $L\infty{}$-norm instead.

\paragraph{L-BFGS Attack}\mbox{}\\
\code{foolbox.attacks.LBFGSAttack}\\
L-BFGS-B is a second-order optimiser that we here use to find the minimum of
\begin{equation*}
    L(\x + \boldsymbol\rho, \ell) + \lambda \left\|\boldsymbol\rho\right\|_2^2\quad\text{s.t.}\quad x_i + \rho_i\in [b_{\text{min}}, b_{\text{max}}]
\end{equation*}
where $\ell\ne\ell_0$ is the target class \citep{szegedy2013}. A line-search is performed over the regularisation parameter $\lambda > 0$ to find the minimum adversarial perturbation. If the target class is not specified we choose $\ell$ as the class of the adversarial example generated by the gradient attack.

\paragraph{SLSQP Attack}\mbox{}\\
\code{foolbox.attacks.SLSQPAttack}\\
Compared to L-BFGS-B, SLSQP allows to additionally specify non-linear constraints. This enables us to skip the line-search and to directly optimise
\begin{equation*}
    \left\|\boldsymbol\rho\right\|_2^2 \quad\text{s.t.}\quad L(\x + \boldsymbol\rho, \ell) = l \,\,\wedge\,\, x_i + \rho_i\in [b_{\text{min}}, b_{\text{max}}]
\end{equation*}
where $\ell\ne\ell_0$ is the target class. If the target class is not specified we choose $\ell$ as the class of the adversarial example generated by the gradient attack.

\paragraph{Jacobian-Based Saliency Map Attack}\mbox{}\\
\code{foolbox.attacks.SaliencyMapAttack}\\
This targeted attack \citep{papernot15} uses the gradient to compute a \textit{saliency score} for each input feature (e.g. pixel). This saliency score reflects how strongly each feature can push the model classification from the reference to the target class. This process is iterated, and in each iteration only the feature with the maximum saliency score is perturbed.

\subsection{Score-Based Attacks}
Score-based attacks do not require gradients of the model, but they expect meaningful scores such as probabilites or logits which can be used to approximate gradients.

\paragraph{Single Pixel Attack}\mbox{}\\
\code{foolbox.attacks.SinglePixelAttack}\\
This attack \citep{localsearch} probes the robustness of a model to changes of single pixels by setting a single pixel to white or black. It repeats this process for every pixel in the image.

\paragraph{Local Search Attack}\mbox{}\\
\code{foolbox.attacks.LocalSearchAttack}\\
This attack \citep{localsearch} measures the model's sensitivity to individual pixels by applying extreme perturbations and observing the effect on the probability of the correct class. It then perturbs the pixels to which the model is most sensitive. It repeats this process until the image is adversarial, searching for additional critical pixels in the neighborhood of previously found ones.

\paragraph{Approximate L-BFGS Attack}\mbox{}\\
\code{foolbox.attacks.ApproximateLBFGSAttack}\\
Same as L-BFGS except that gradients are computed numerically. Note that this attack is only suitable if the input dimensionality is small.

\subsection{Decision-Based Attacks}
Decision-based attacks rely only on the class decision of the model. They do not require gradients or probabilities.

\paragraph{Boundary Attack}\mbox{}\\
\code{foolbox.attacks.BoundaryAttack}\\
Foolbox provides the reference implementation for the Boundary Attack~\citep{boundaryattack}. The Boundary Attack is the most effective decision-based adversarial attack to minimize the L2-norm of adversarial perturbations. It finds adversarial perturbations as small as the best gradient-based attacks without relying on gradients or probabilities.

\paragraph{Pointwise Attack}\mbox{}\\
\code{foolbox.attacks.PointwiseAttack}\\
Foolbox provides the reference implementation for the Pointwise Attack. The Pointwise Attack is the most effective decision-based adversarial attack to minimize the L0-norm of adversarial perturbations.

\paragraph{Additive Uniform Noise Attack}\mbox{}\\
\code{foolbox.attacks.AdditiveUniformNoiseAttack}\\
This attack probes the robustness of a model to i.i.d. uniform noise. A line-search is performed internally to find minimal adversarial perturbations.

\paragraph{Additive Gaussian Noise Attack}\mbox{}\\
\code{foolbox.attacks.AdditiveGaussianNoiseAttack}\\
This attack probes the robustness of a model to i.i.d. normal noise. A line-search is performed internally to find minimal adversarial perturbations.

\paragraph{Salt and Pepper Noise Attack}\mbox{}\\
\code{foolbox.attacks.SaltAndPepperNoiseAttack}\\
This attack probes the robustness of a model to i.i.d. salt-and-pepper noise. A line-search is performed internally to find minimal adversarial perturbations.

\paragraph{Contrast Reduction Attack}\mbox{}\\
\code{foolbox.attacks.ContrastReductionAttack}\\
This attack probes the robustness of a model to contrast reduction. A line-search is performed internally to find minimal adversarial perturbations.

\paragraph{Gaussian Blur Attack}\mbox{}\\
\code{foolbox.attacks.GaussianBlurAttack}\\
This attack probes the robustness of a model to Gaussian blur. A line-search is performed internally to find minimal blur needed to turn the image into an adversarial.

\paragraph{Precomputed Images Attack}\mbox{}\\
\code{foolbox.attacks.PrecomputedImagesAttack}\\
Special attack that is initialized with a set of expected input images and corresponding adversarial candidates. When applied to an image, it tests the models robustness to the precomputed adversarial candidate corresponding to the given image. This can be useful to test a models robustness against image perturbations created using an external method.

\section{Acknowledgements}
\label{acknowledgements}

This work was supported by the Carl Zeiss Foundation (0563-2.8/558/3), the Bosch Forschungsstiftung (Stifterverband, T113/30057/17), the International Max Planck Research School for Intelligent Systems (IMPRS-IS), the German Research Foundation (DFG, CRC 1233, Robust Vision: Inference Principles and Neural Mechanisms) and the Intelligence Advanced Research Projects Activity (IARPA) via Department of Interior/Interior Business Center (DoI/IBC) contract number D16PC00003. The U.S. Government is authorized to reproduce and distribute reprints for Governmental purposes notwithstanding any copyright annotation thereon. Disclaimer: The views and conclusions contained herein are those of the authors and should not be interpreted as necessarily representing the official policies or endorsements, either expressed or implied, of IARPA, DoI/IBC, or the U.S. Government.

\bibliography{references}
\bibliographystyle{icml2017}

\end{document}